\documentclass{article}
\usepackage{ijcai25}
\usepackage{times}
\usepackage{soul}
\usepackage{url}
\usepackage[hidelinks]{hyperref}
\usepackage[utf8]{inputenc}
\usepackage[small]{caption}
\usepackage{graphicx}
\usepackage{amsmath}
\usepackage{amsthm}
\usepackage{booktabs}
\usepackage{algorithm}
\usepackage{algorithmic}
\usepackage[switch]{lineno}

\usepackage{cleveref}
\usepackage{bm}
\usepackage[accsupp]{axessibility} %
\usepackage{epsfig}
\usepackage{amssymb}
\usepackage{enumitem}
\usepackage{multirow}
\usepackage{xspace}
\usepackage{arydshln}  %
\usepackage[table,dvipsnames]{xcolor}  %
\usepackage{soul}  %
\usepackage[most]{tcolorbox}
\usepackage{pifont}
\usepackage{nicematrix}  %
\usepackage{makecell}
\usepackage{subcaption}

\usepackage{booktabs}    % 提供 \toprule, \midrule, \bottomrule
\usepackage{colortbl}    % 提供 \rowcolor
\usepackage[table]{xcolor} % 更强大的颜色支持，推荐代替 colortbl
\usepackage{arydshln}    % 提供 \cdashline
\usepackage{caption}     % 用于调整表格和图表的标题格式（如果需要）

\newcommand{\tablestyle}[2]{\setlength{\tabcolsep}{#1}\renewcommand{\arraystretch}{#2}\centering\small}
\newcommand{\avgn}{Avg$^\text{N}$\xspace}
\newcommand{\mmbdev}{MMB\xspace}
\newcommand{\mmbdevcn}{MMB$^{\text{cn}}$\xspace}
\newcommand{\mmep}{MME$^{\text{P}}$\xspace}
\newcommand{\seedimg}{SEED$^{\text{I}}$\xspace}
\newcommand{\pope}{POPE\xspace}

\newcommand{\ian}{IAN\xspace}

\title{
  MAGE:  Multimodal Alignment and Generation Enhancement via Bridging Visual and Semantic Spaces
}

\author{
Shaojun E$^{1,3, *}$\and
Yuchen Yang$^{2,*}$\and
Jiaheng Wu$^2$\and
Yan Zhang$^{1}$\and
Tiejun Zhao$^{2}$\and
Ziyan Chen$^{1,\dagger}$\\
\affiliations
$^{1}$ Global Tone Communication Technology Co., Ltd., Beijing, China\quad \\
$^{2}$ Faculty of computing, Harbin Institute of Technology, Harbin, China\quad \\
$^{3}$ School of Computer Science and Technology, Beijing Jiaotong University, Beijing, China
\emails
\{eshaojun, zhangyan01, chenziyan\}@gtcom.com.cn,\\
\{23S003057, 23S136126\}@stu.hit.edu.cn; \
tjzhao@hit.edu.cn
}

\begin{document}
\maketitle
% \let\thefootnote\relax
% \footnotemark\footnotetext{$^\dagger$ Equal contribution, $^*$ Corresponding author.}
\begin{abstract}
In the latest advancements in multimodal learning, effectively addressing the spatial and semantic losses of visual data after encoding remains a critical challenge. This is because the performance of large multimodal models is positively correlated with the coupling between visual encoders and large language models. Existing approaches often face issues such as vector gaps or semantic disparities, resulting in information loss during the propagation process. To address these issues, we propose \textbf{\it{MAGE}} (\textbf{M}ultimodal \textbf{A}lignment and \textbf{G}eneration \textbf{E}nhancement), a novel framework that bridges the semantic spaces of vision and text through an innovative alignment mechanism. By introducing the Intelligent Alignment Network (IAN), \textbf{\it{MAGE}} achieves dimensional and semantic alignment. To reduce the gap between synonymous heterogeneous data, we employ a training strategy that combines cross-entropy and mean squared error, significantly enhancing the alignment effect. Moreover, to enhance MAGE’s “Any-to-Any” capability, we developed a fine-tuning dataset for multimodal tool-calling instructions to expand the model’s output capability boundaries. Finally, our proposed multimodal large model architecture, MAGE, achieved significantly better performance compared to similar works across various evaluation benchmarks, including MME, MMBench, and SEED. Complete code and appendix are available at: \url{https://github.com/GTCOM-NLP/MAGE}
\end{abstract}

\section{Introduction}
\label{sec:intro}

\begin{figure*}[!t] 
    \begin{center}
        \includegraphics[width=0.85\textwidth]{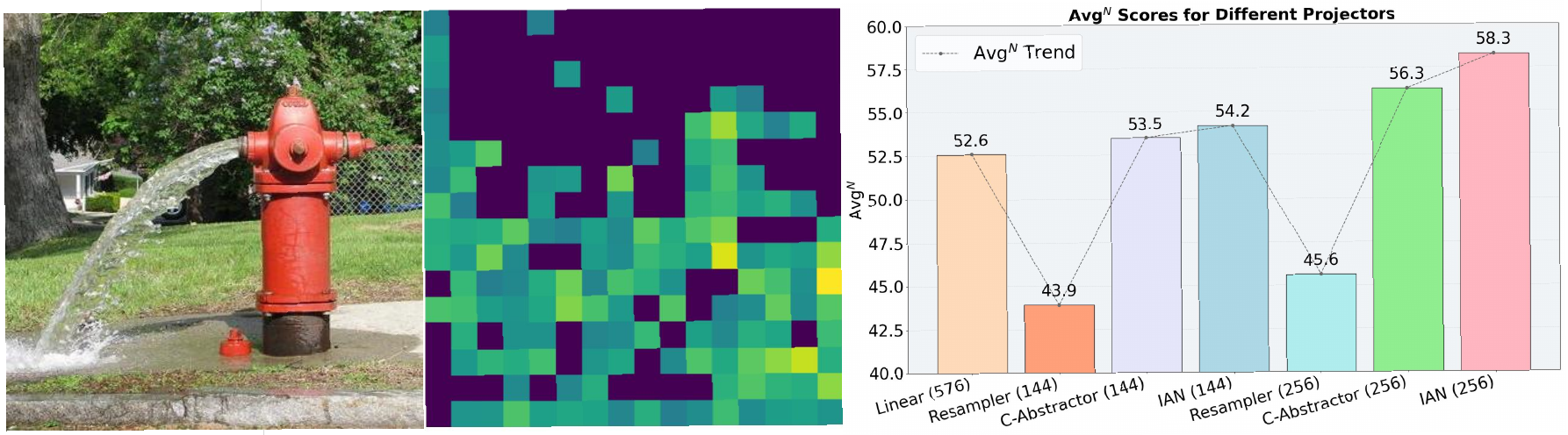}
    \end{center}
    \caption{
        (Left) an example of an attention map from the IAN. (middle) The attention matrix of a visual token obtained after passing through IAN in the left figure.and (right) A comparison of the spatial understanding capability between IAN and common projectors, where $Avg^N$ represents the average performance computed from six common understanding tasks, including MME, MMB, and SEED.
    }
    \label{fig:every projector’s performance}
\end{figure*}

In recent years, with the remarkable progress of large language models (LLMs) in natural language processing tasks \cite{brown2020language,vaswani2017attention}, researchers have gradually extended their powerful capabilities to multimodal domains, enabling these models to simultaneously comprehend and generate both visual and textual information \cite{anthropic2024claude3d5,team2023gemini,openai2024hellogpt4o}. \cref{fig:every projector’s performance} shows the performance of common projection mechanisms. However, achieving effective integration of visual encoders and LLMs still faces significant challenges in cross-modal alignment, including semantic gaps and dimensional mismatches \cite{yu2022coca,yuan2021florence}. These challenges directly impact the performance and efficiency of the models in multimodal tasks, limiting the practical applications of multimodal large language models (MLLMs).

Existing methods for aligning visual and linguistic modalities face several limitations \cite{dai2024nvlm}. First, linear projectors preserve the integrity of visual features through simple mappings but struggle to model semantic relationships between modalities, often resulting in semantic information loss in complex tasks. Second, projector-based mechanisms, such as Resamplers \cite{flamingo} and Q-Formers \cite{blip2-li2023blip}, improve alignment through query mechanisms and cross-modal attention but overlook the global semantic information of visual features, leading to suboptimal performance in deep semantic tasks. Finally, recent approaches like TokenPacker \cite{tokenpacker-li2024tokenpacker} and Honeybee \cite{honeybee-cha2024honeybee} have made progress by enhancing efficiency and preserving local features, yet they insufficiently address the role of loss functions in alignment. Most methods rely on single-generation or alignment losses, failing to fully capture deep semantic relationships between visual and linguistic modalities \cite{wang2024qwen2,instructblip-dai2023instructblip}.

\begin{figure*}[!t] 
    \begin{center}
        \includegraphics[width=\textwidth]{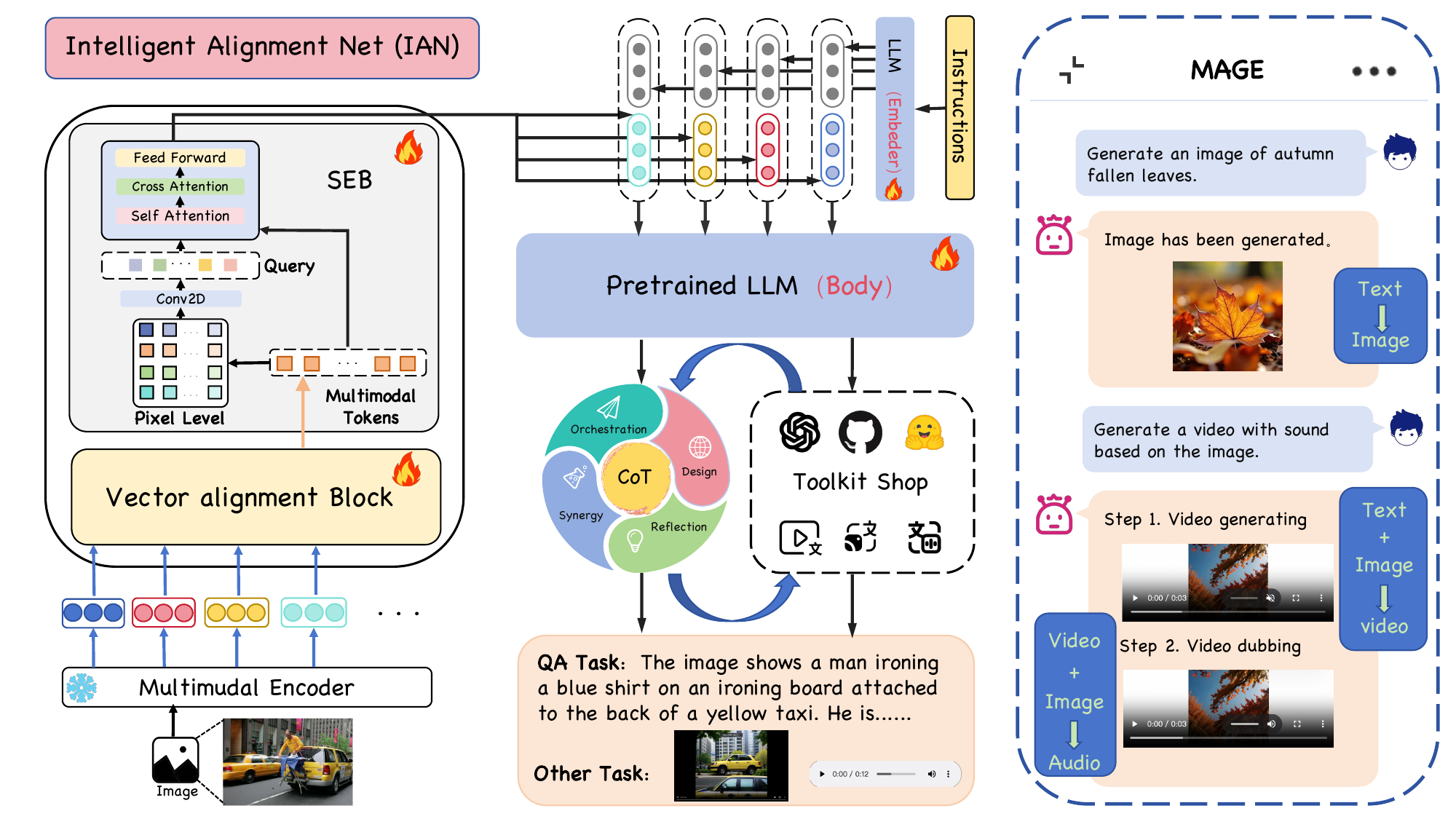}
    \end{center}
    \caption{
        \textbf{The detailed aspects of MAGE.}
        % This figure provides a detailed illustration of the specific implementation details of MAGE. IAN consists of two components: vector alignment and semantic enhancement. The aligned vectors after encoding encapsulate rich semantic knowledge. The trained LLM is capable of decomposing and planning instructions, ultimately reorganizing them into different subtasks to schedule various tool models, thereby fulfilling user instructions.
        IAN consists of two components: vector alignment and semantic enhancement. The encoded vectors encapsulate rich semantic knowledge, enabling the LLM to decompose and plan instructions into different subtasks, thereby scheduling various tool models to fulfill user instructions.
    }
    \label{fig:detail_framework}
\end{figure*}

The limitations of these approaches indicate that existing research has yet to fully address the dual challenges of dimensional consistency and semantic integrity in the alignment of visual and linguistic modalities. Therefore, there is an urgent need for a solution that simultaneously optimizes both efficiency and semantic alignment.

To address the challenges of semantic and dimensional gaps between visual encoders and large language models (LLMs), while simultaneously meeting the requirements for generation and alignment in multimodal tasks, we propose a novel multimodal large model architecture, named MAGE, as illustrated in \cref{fig:detail_framework} . MAGE aims to achieve efficient and flexible cross-modal generation and processing through innovative designs in three aspects: semantic alignment, training strategy, and interaction framework.

The core architecture of MAGE comprises three innovative modules. First, we introduce the Intelligent Alignment Network module, which consists of two submodules: the Vector Alignment Module and the Semantic Enhancement Module. The Vector Alignment Module maps visual feature vectors to a dimensional space consistent with the LLM input, ensuring structural dimensional alignment and eliminating modality gaps. After alignment, the Semantic Enhancement Module injects high-level semantic information into the visual features, thereby enhancing their expressive power in the semantic space of the language model and providing strong support for cross-modal tasks.

Second, to further improve the effectiveness of cross-modal alignment, we propose a dual-loss training strategy. This strategy includes the Image-Text Generation loss and the Image-Text Distance Minimization loss. The ITG loss guides the model to generate natural language descriptions from visual inputs, enhancing cross-modal alignment from the perspective of generative tasks. Meanwhile, the ITDM loss minimizes the semantic distance between visual and textual features, further reinforcing deep semantic consistency between the two modalities in alignment tasks. These two loss components complement each other, effectively improving the model's performance in both generation and alignment tasks.

Finally, to adapt to complex task scenarios, we expand the output end into a scheduling terminal supporting multiple agents and construct a multi-task scheduling dataset. Through training, MAGE can understand user instructions, perform task decomposition and planning, and call different tool models to achieve the goal. Unlike traditional agent systems driven by large language models (e.g., HuggingGPT), MAGE is a multi-tool invocation model driven by multimodal large language model. Its advantage lies in integrating multiple modalities, enabling more accurate understanding and generation of cross-modal task instructions. At the same time, through flexible tool invocation, it achieves more efficient task execution and supports richer application scenarios.

In summary, the proposed MAGE achieves significant performance improvements in multiple multimodal benchmarks by employing a superior semantic alignment strategy, surpassing existing methods. Experimental results demonstrate that even with a reduction in visual tokens, MAGE outperforms current methods in task performance, resource efficiency, and output flexibility.

The main contributions of this work are summarized as follows:

\begin{itemize}[leftmargin=.3cm,noitemsep,nosep]
    \item 
        We proposed a novel cross-modal alignment architecture, IAN (Intelligent Alignment Network), which integrates a Vector Alignment Module and a Semantic Enhancement Module. Experiments show that this architecture effectively addresses the challenges of dimensional mismatch and semantic gaps between visual and linguistic modalities.
    \item
        A dual-loss training strategy was designed, comprising the Image-Text Generation loss and the Image-Text Distance Minimization loss, enabling end-to-end optimization of the semantic alignment process.
    \item 
        % We constructed a multimodal tool-scheduling dataset HMDSet(human-machine dialogue dataset) and developed an innovative multimodal agent framework. This framework dynamically organizes structured configuration information to accomplish composite generation tasks and supports a wide range of multimodal applications, including image generation, audio generation, video generation, and image editing, significantly extending the model's capability boundaries.
        We constructed a multimodal tool-scheduling dataset, HMDSet (a dialogue dataset), and developed an innovative framework. This framework dynamically organizes structured configuration information to accomplish composite generation tasks, significantly extending the model's capability boundaries.
\end{itemize}

\section{Related Work}
\label{sec:related_work}

\subsection{Multimodal Large Language Models}
\label{rel:mllm}

In recent years,
multimodal large models such as Flamingo \cite{flamingo}, BLIP-2 \cite{blip2-li2023blip}, LLaVA \cite{llava-liu2024visual}, and their subsequent versions (e.g., LLaVA-v1.5 \cite{llava1d5-liu2024improved} and LLaVA-next \cite{llavanext-li2024llava}) have achieved significant progress. Flamingo integrates visual features with language models through the introduction of the Perceiver Resampler and cross-modal attention mechanisms, demonstrating exceptional performance in few-shot learning scenarios. BLIP-2 employs a lightweight Q-Former module to bridge the modality gap between visual encoders and frozen LLMs, thereby improving zero-shot image-to-text generation capabilities. The LLaVA series further enhances multimodal capabilities in vision-language tasks through innovative cross-modal alignment strategies and large-scale image-text pretraining.

\subsection{Visual Projector in VLMs}
\label{rel:proj}
In multimodal large language models (VLMs), the visual projection layer is a critical module for achieving cross-modal alignment between visual information and language models.
classical projection layers can generally be categorized into two types: MLP-based projectors and query transformer (Q-Former)-based projectors \cite{blip2-li2023blip,instructblip-dai2023instructblip}. 

In recent years, emerging methods \cite{honeybee-cha2024honeybee,tokenpacker-li2024tokenpacker,zhao2024easygen} have aimed to strike a balance between efficiency and performance. However, existing approaches fail to adequately consider semantic integrity and alignment precision during the alignment process, which provides a strong motivation to explore more efficient and precise visual-language alignment strategies.

\subsection{Any-to-Any Model}
Recent efforts aim to build models capable of arbitrary modality transformation, mimicking human flexibility. Models like NExT-GPT and CoDi show promise, but struggle with quality and precision in complex tasks. Agent systems such as HuggingGPT \cite{hugginggpt-shen2024hugginggpt} and MetaGPT \cite{metagpt-hong2023metagpt} integrate tools with LLMs to support dynamic multimodal tasks, yet still depend on separate models for understanding and generation—limiting their ability to handle implicit or complex instructions. LLaVA-PLUS \cite{llavaplus-liu2025llava} aims to enhance multimodal language understanding, focusing on improving LLMs’ ability in language comprehension through the Agent approach. However, it fails to fully address the scalability issue of agents in handling multimodal outputs.

\section{Method}
\label{sec:method}
% In contemporary multimodal learning research, effectively aligning visual and textual information, along with extending the model's output modalities and enhancing its generative capabilities, are pivotal for developing more efficient Multimodal Large Language Models.To address this challenge, we propose an Intelligent Alignment Network (IAN) module to enhance visual-textual alignment in multimodal learning by bridging visual encoder outputs to the semantic space of large language models, reducing semantic loss. Our innovative training strategy uses image-text pair alignment with specialized loss functions, optimizing semantic alignment and improving multimodal task performance. Additionally, we extend the model’s output capabilities by integrating multimodal tools, enabling both textual and multimodal outputs, thus enhancing its flexibility and broadening future application potential.
% In multimodal learning, aligning visual and textual information, expanding output modalities, and enhancing generative capabilities are key to improving the efficiency of multimodal large language models. To address this, we propose the Intelligent Alignment Network (IAN), which bridges visual encoders and the semantic space of large language models, reducing semantic loss and strengthening alignment. By combining image-text alignment with specialized loss functions, task performance is optimized, while the integration of multimodal tools extends output capabilities, enabling both textual and multimodal generation, enhancing flexibility and application potential.

\subsection{IAN: an Intelligent Alignment Network}

The architecture of IAN consists of two core components: the VAB (Vector Alignment Block) and the SEB (Semantic Enhancement Block). These two modules were introduced to address key limitations in prior research, where approaches often relied on simplistic strategies like a single linear layer or a Q-former initialized with random learnable parameters \cite{llava-liu2024visual,blip2-li2023blip,honeybee-cha2024honeybee,yao2024deco}. Such methods struggled to achieve effective semantic and dimensional alignment simultaneously.

Based on previous research, we found that reducing visual tokens doesn't require complex visual-semantic extractors like Q-former \cite{yao2024deco}. Simple 2D pooling operations at the patch level can effectively reduce tokens while maintaining strong performance. Building on this, we proposed replacing fixed-parameter pooling layers with learnable convolutional layers to enable adaptive learning of mapping patterns. This idea is key to SEB, where trainable convolutional layers refine and enhance visual embeddings to better align with the language model. To address challenges in dimensional alignment and semantic enhancement, we designed IAN with two modules: VAB for alignment and SEB for enhancement. Unlike previous methods that used a single linear layer or Q-former for both tasks—often oversimplifying or complicating the process—our modular approach decouples these functions, allowing VAB to align dimensions and SEB to refine semantic representations more effectively.

\paragraph{VAB Module} We design a flexible linear projection network combined with multilayer nonlinear transformations, enabling the efficient alignment of high-dimensional embeddings from the multimodal encoder with the low-dimensional embedding space of the language model while preserving the original semantic contextual information of the multimodal encoder. Specifically, VAB receives embeddings $\mathbf{V} \in \mathbb{R}^{N \times D_v}$ from the multimodal encoder, where $N$ represents the number of visual features and $D_v$ denotes the embedding dimension of the multimodal encoder. These embeddings undergo nonlinear transformations through MLP and learnable normalization layers, ultimately producing aligned embeddings $\mathbf{V}' \in \mathbb{R}^{N \times D_l}$, where $D_l$ is the embedding dimension of the language model.

\paragraph{SEB Module} We propose the SEB to enhance semantic consistency between visual embeddings and the language model via cross-modal self-attention. The image is processed by the CLIP \cite{radford2021learning} and aligned by VAB to generate a global visual embedding \(\mathbf{a} \in \mathbb{R}^{N \times D_l}\), capturing high-level features. To introduce local detail, the image is divided into patches and passed through a CNN \cite{wu2017introduction} to obtain a new embedding \(\mathbf{b} \in \mathbb{R}^{N \times D_b}\),  which serves as the query.In the attention mechanism, \(\mathbf{b}\) acts as the query, while \(\mathbf{a}\) is used as both key and value. The attention computation integrates visual and semantic information, resulting in a refined embedding. The output embedding \(\mathbf{V}'' \in \mathbb{R}^{N \times D_l}\) incorporates both global and local visual features, effectively bridging the gap between the visual encoder and the large language model.

\subsection{Aligning IAN to LLM}

% In visual language models (VLMs), effective alignment between images and text is essential for improving performance on multimodal tasks and enhancing semantic understanding. While existing methods use cross-entropy loss for model optimization, this approach has limitations that can restrict the model’s multimodal capabilities. We propose a new alignment strategy that combines cross-entropy loss with mean squared error (MSE) loss to improve the alignment between visual encodings and language embeddings.
In visual language models (VLMs), effective alignment between images and text is essential for enhancing performance on multimodal tasks and improving semantic understanding. While existing methods typically use cross-entropy loss for optimization, this approach has limitations that restrict the model’s multimodal capabilities. We propose combining cross-entropy loss with mean squared error (MSE) loss to better align visual encodings and language embeddings.
\subsubsection{Motivation for the New Alignment Strategy}

Current methods predominantly rely on cross-entropy loss to optimize language models for accurate text generation \cite{llava-liu2024visual,blip2-li2023blip,honeybee-cha2024honeybee,tokenpacker-li2024tokenpacker}. However, this approach does not directly minimize the distance between visual and language embeddings, leading to weak alignment and suboptimal performance in multimodal tasks. Furthermore, cross-entropy loss lacks explicit guidance on the relationship between image and text embeddings, which can result in inconsistent representations. To address these limitations, we propose combining Mean Squared Error (MSE) loss with cross-entropy loss, where once the image features are mapped by the projection layer, they should align with the image caption encoded by the large language model \cite{zhao2024easygen}. Additionally, we introduce an image-text distance minimization (ITDM) loss alongside the ITG loss to directly minimize the distance between visual and text embeddings, thereby enhancing both language generation and multimodal alignment.

\subsubsection{Specific Alignment Strategy}

Based on the above motivations, our proposed alignment strategy employs two primary loss functions: Image-Guided Text Generation (ITG) loss and Image-Text Distance Minimization (ITDM) loss.

The ITG loss is designed to guide the model in generating corresponding text based on the input image. Specifically, the input image is first processed by the CLIP model to obtain its representation vector. This vector is then passed through the IAN module, where it is mapped into the text embedding space of the large language model. Subsequently, the ITG loss is computed in an autoregressive manner. The loss function can be expressed as follows:
\begin{equation}
    \mathcal{L}_{\mathrm{ITG}} = -\frac{1}{L} \sum_{i=1}^{L} \log p_{\theta}(y_{i} | y_{<i}, \mathbf{X}_{\mathrm{img}}, \mathbf{X}_{\mathrm{text}}),
    \label{eq:itg_loss}
\end{equation}
where $\mathbf{}{Y} = \{y_i\}_{i=1}^{L}$ denotes the aligned visual embedding produced by IAN, $\mathbf{X}_{\mathrm{img}}$ denotes the image input, while $\mathbf{X}_{\mathrm{text}}$ represents the textual instructions. The model parameters are represented by $\theta$, which also include the parameters of the projection layer, denoted as IAN.

The ITDM loss function aims to minimize the distance between the image vector processed by IAN and the text vector encoded by the LLM. Specifically, the image vector processed by IAN is represented as $\mathbf{d}_{\mathrm{ian}}$, and the text vector encoded by the LLM is represented as $\mathbf{d}_{\mathrm{llm}}$. The ITDM loss is calculated using the following formula:
\begin{equation}
    \mathcal{L}_{\mathrm{ITDM}} = \frac{1}{N} \sum_{i=1}^{N} \| \mathbf{d}_{\mathrm{ian}} - \mathbf{d}_{\mathrm{llm}} \|_2^2,
    \label{eq:itdm_loss}
\end{equation}
Here, $N$ represents the batch size.
To fully leverage the advantages of both the cross-entropy loss and the mean squared error loss, we define a composite loss function as follows:
\begin{equation}
    \mathcal{L} = \mathcal{L}_{\mathrm{ITG}} + \mathcal{L}_{\mathrm{ITDM}},
    \label{eq:combined_loss}
\end{equation}
The composite loss function improves language generation and aligns visual and language embeddings. Through contrastive learning, multimodal vectors are aligned with the semantic space of large language models. After fine-tuning, the model better processes multimodal information and enhances its outputs.
\begin{figure*}[!t]
    \begin{center}
    \scalebox{0.8}{
        \includegraphics[width=\textwidth]{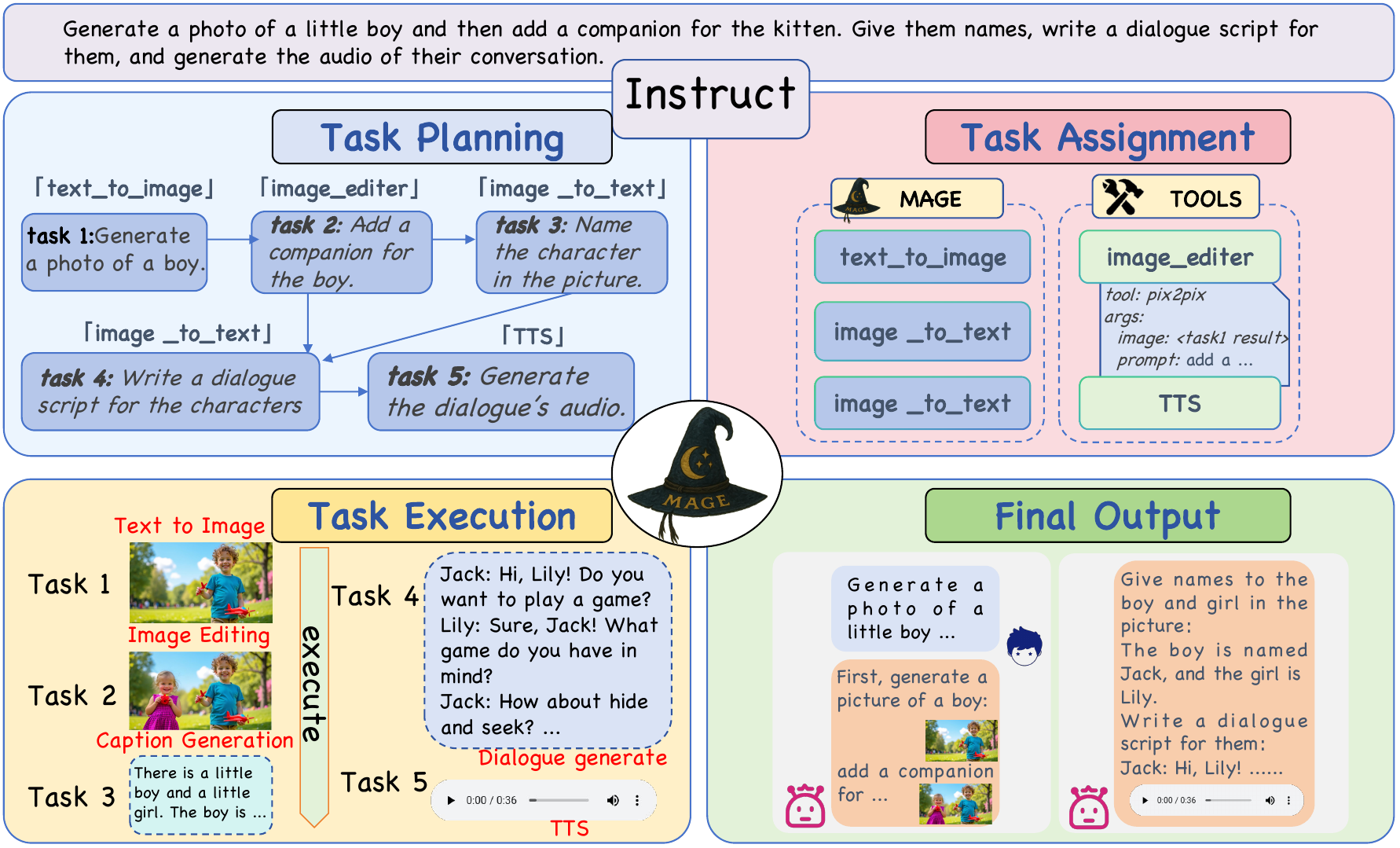}
    }
    \end{center}
    \caption{
        \textbf{Hierarchical Workflow of the Proposed Agent Framework.}This figure illustrates a hierarchical process for generating an image of a boy with a companion. The process includes task planning, task assignment, task execution, and final output.).
    }
    \label{fig:Agent_Forecast_Params}
\end{figure*}

\subsection{Extended SKILLS with Multimodal Tool Use}

% In traditional vision-language models, outputs are often limited to textual descriptions, such as image captioning or question answering. This single-modal output approach has significant limitations in real-world applications, particularly in scenarios requiring rich multimodal expressions, such as image generation, dynamic video generation, or multimodal interactive tasks. To address this issue, we propose a novel architectural design that supports multimodal outputs, equipping the model with enhanced task planning capabilities and tool utilization abilities to meet complex and diverse user demands.
To expand the input boundaries of the model, we propose a novel architectural design that supports multimodal outputs and enhances the model’s task planning capabilities to solve complex tasks.

\paragraph{HMDSet Dataset Construction}
% To enable MAGE to achieve complex cross-modal understanding and content generation through tool utilization, we designed a human-machine dialogue dataset(HMDSet) that includes a wide range of multimodal inputs and outputs. Specifically, the user instructions in the dataset contain at least one textual modality and can encompass various modalities such as text, images, audio, and video. The output of MAGE is in the form of a JSON task planning result, which includes the tool categories and tool parameters.
We designed a human-machine dialogue dataset (HMDSet) that includes a wide range of input and output modalities, covering text, images, audio, video, and complex task handling. This enables MAGE to tackle complex cross-modal understanding and generation tasks.
\paragraph{Hierarchical Architecture and Output Format of the Agent}
The unique aspect of this multimodal agent lies in its ability to call the required toolset in a structured JSON format based on natural language output generated by a large language model (LLM), as shown in \cref{fig:Agent_Forecast_Params}. We employ a hierarchical strategy in the task planning and prediction process to efficiently determine the specific tools and parameters.

\renewcommand{\arraystretch}{0.8}
\begin{table*}[!ht]
    \tablestyle{3pt}{1.12}
    \centering
    \scalebox{0.9}{
    \begin{tabular}{l|ccccc|ccccc}
        \toprule
        {Method} & {LLM} & {Vision Encoder} & {Projector} & {\#Token} & {Res.} & \mmbdev & \mmbdevcn & \seedimg & {MME} & \pope \\

        \hline\hline
        \rowcolor[gray]{0.85}\multicolumn{11}{l}{\textit{\textbf{Approaches using 7B LLM}}} \\ 
        \hline        
        Qwen-VL-Chat \cite{qwenvl-bai2023qwen}        &Qwen-7B &OpenCLIP ViT-bigG & Resampler & 256 &448 &60.6 &56.3 &58.2 & 1487/\textbf{360} & - \\
        InstructBLIP \cite{instructblip-dai2023instructblip}         &Vicuna-7B &EVA-CLIP ViT-G & Q-former & 64 &224 &36.0 &23.7 &53.4 & - & - \\
        LLaVA-TokenPacker-HD \cite{tokenpacker-li2024tokenpacker}    & Vicuna-7B & CLIP ViT-L/14 & TokenPacker & 64 & 336 & 67.4 & - & - & 1489/338 & 86.3 \\
        LLaVA-1.5 \cite{llava1d5-liu2024improved} &Vicuna-7B & CLIP ViT-L/14 & Linear & 576 & 336 & 64.3 & 58.3 & 58.6 & 1510/- & 85.9 \\
        LLaVA-NeXT \cite{llavanext-li2024llava} &Vicuna-7B &CLIP ViT-L/14 & Linear & 576 & 336 & 67.4 & 62.3 & - & 1519/332 & 86.5 \\
        C-Abstractor \cite{honeybee-cha2024honeybee} & Vicuna-7B & CLIP ViT-L/14 & C-Abstractor & 144 & 224 & 70.1 & - & 64.5 & \textbf{1584}/307 & - \\
        D-Abstractor \cite{honeybee-cha2024honeybee} & Vicuna-7B & CLIP ViT-L/14 & D-Abstractor & 144 & 224 & 70.8 & - & 63.8 & 1544/291 & - \\
        \cdashline{1-11}

        MAGE(Ours) & Vicuna-7B & CLIP ViT-L/14 & IAN & 144 & 336 & \textbf{71.2} & \textbf{70.3} & \textbf{65.3} & 1572/323 & \textbf{86.7} \\
        \midrule\hline
        \rowcolor[gray]{0.85}\multicolumn{11}{l}{\textit{\textbf{Approaches using 13B LLM}}} \\ 
        \hline
        InstructBLIP \cite{instructblip-dai2023instructblip}       &Vicuna-13B &EVA-CLIP ViT-G & Q-former & 64 &224 & - & - & - & 1212/- & 78.9 \\
        LLaVA-TokenPacker-HD \cite{tokenpacker-li2024tokenpacker}    & Vicuna-13B & CLIP ViT-L/14 & TokenPacker & 64 & 336 & 69.5 & - & - & 1595/\textbf{356} & 88.1 \\
        LLaVA-1.5 \cite{llava1d5-liu2024improved} & Vicuna-13B & CLIP ViT-L/14 & Linear & 576 & 336 & 67.7 & 63.6 & 61.6 & 1531/- & 85.9 \\
        LLaVA-NeXT \cite{llavanext-li2024llava} &Vicuna-13B &CLIP ViT-L/14 & Linear & 576 & 336 & 70 & 68.5 & - & 1575/326 & 86.2 \\
        C-Abstractor \cite{honeybee-cha2024honeybee} & Vicuna-13B & CLIP ViT-L/14 & C-Abstractor & 256 & 336 & 73.2 & - & \textbf{68.2} & 1629/315 & - \\
        D-Abstractor \cite{honeybee-cha2024honeybee} & Vicuna-13B & CLIP ViT-L/14 & D-Abstractor & 256 & 336 & 73.5 & - & 66.6 & \textbf{1632}/333 & - \\
        \cdashline{1-11}
        MAGE(Ours) & Vicuna-13B & CLIP ViT-L/14 & IAN & 144 & 336 & \textbf{73.9} & \textbf{73.0} & 67.8 & \textbf{1632}/312 & \textbf{89.9} \\
        \bottomrule
    \end{tabular}
    }
    \caption{\textbf{Comparison with other SOTA.}Res means the resolution of the image encoder, and \#Token represents the number of visual tokens. The best results are shown in bold, and the second-best are underlined. }
    \label{tab:comp_others}
\end{table*}
\renewcommand{\arraystretch}{1.0}
\section{Experiments}
\label{sec:experiments}

\subsection{Implementation Details}
In this section, we describe the experimental setup used in our study to evaluate the performance of our proposed model. The following outlines the implementation details, datasets, and benchmarks that were used to ensure robustness and comprehensiveness of the results.
\paragraph{Model Configuration.}
We used two variants of the Vicuna-v1.5 \cite{vicuna1d5-zheng2023judging} large language model (7B and 13B) and CLIP ViT-L/14 \cite{clip-radford2021learning} (336px) as the vision encoder, with input images set to a resolution of 336×336 pixels. Visual features are derived from the penultimate layer of the CLIP model, excluding the CLS token. During training, we fine-tuned all model parameters (including CLIP and LLM) without using parameter-efficient techniques like LoRA \cite{lora-hu2021lora}, ensuring the model learns complete representations. Training was conducted using 48 A6000 GPUs. All the training data is presented in \cref{table:datasets}.Our training process is divided into three phases:
\begin{table}[!t]
    \tablestyle{2pt}{1.05}
    \begin{center}
    \scalebox{1}{
        \begin{tabular}{@{}l|p{4.2cm}|c@{}}
        \toprule
        Task & Datasets  & \#samples  \\
        \midrule
        Pretraining dataset & BlipCapFilt\cite{li2022blip} & 558K \\ 
        General Instrution & LLaVA150K\cite{llava-liu2024visual}, ShareGPT\cite{vicuna2023}&  665K  \\
        Doc/Chart/Screen    & 
        Doc-VQA\cite{mathew2021docvqa}, ChartQA\cite{masry2022chartqa}, DVQA\cite{kafle2018dvqa} &  43K \\
        Math/Reasoning  & GeoQA+\cite{chen2021geoqa},    & 17K   \\
        General OCR & SynthDog-EN\cite{kim2022ocr}          & 40K   \\
        Tool Use  & HMDSet    & 31K   \\
        \bottomrule
        \end{tabular}
    }
    \end{center}
    \caption{
        List of all training datasets. 
    }
    \label{table:datasets}
\end{table}
% vicuna:\bibitem[Chiang et~al.(2023)Chiang, Li, Lin, Sheng, Wu, Zhang, Zheng, Zhuang,
%   Zhuang, Gonzalez, Stoica, and Xing]{vicuna}
% Wei-Lin Chiang, Zhuohan Li, Zi Lin, Ying Sheng, Zhanghao Wu, Hao Zhang, Lianmin
%   Zheng, Siyuan Zhuang, Yonghao Zhuang, Joseph~E. Gonzalez, Ion Stoica, and
%   Eric~P. Xing.
% \newblock {Vicuna: An Open-Source Chatbot Impressing GPT-4 with 90\%* ChatGPT
%   Quality}, 2023.

\paragraph{Stage\,\,1:Pretraining Details.}
For the initial training phase, we use the CC-558K dataset, to perform semantic alignment across modalities. In this phase, we freeze the parameters of the visual encoder (CLIP) and the LLM, focusing solely on training the IAN. This approach enables us to align the visual and linguistic features without altering the core model components.
\paragraph{Stage\,\,2:Instruction-tuning Details.}
In the second phase, we scale up the dataset to approximately 980K samples by combining the 665K mixture dataset with the ALLaVA-Instruct-VFLAN-4V \cite{allava-chen2024allava} dataset. This extended dataset includes a rich set of question-answer pairs, simulating real-world user interactions. For training in this phase, only 90\% of this dataset was used. This phase enables fine-tuning of the model for more natural and contextually relevant multimodal responses.
\paragraph{Stage\,\,3:Self-Cognition Training Details}
We refer to the third stage of training as "Self-Awareness Planning Training," aimed at enhancing the model's ability to handle complex cross-modal understanding and generation. The training data combines the remaining 10\% of the fine-tuning dataset from the second stage with the newly constructed self-awareness dataset, which includes multimodal human-machine dialogues involving text, images, audio, and video. The output is a JSON task plan specifying tool categories and parameters. The dialogue examples are generated by GPT-4 and manually reviewed to ensure diversity and accuracy. This stage of training enables the model to develop a chain-of-thought capability, allowing it to proactively invoke tools and autonomously plan and execute complex tasks.
\subsection{Experimental Setting}
We introduce experimental settings, including the benchmarks and evaluation metrics.
\paragraph{Benchmarks.}
We evaluate the model's performance using four multimodal large language model (MLLM) benchmarks, each targeting different aspects of multimodal understanding. MME \cite{mmedataset-fu2024mmecomprehensiveevaluationbenchmark} assesses perception and cognition capabilities using a binary yes/no format; MMBench \cite{mmbench-liu2025mmbench} evaluates reasoning and multimodal tasks through multiple-choice questions; SEED-Bench \cite{seedbench-li2023seed} tests the model’s ability to handle complex multimodal queries; and POPE \cite{pope-li2023evaluating} detects model hallucinations using binary questions to assess the accuracy of providing correct information.
\paragraph{Evaluation Metrics.}We use the official evaluation scripts provided for each benchmark to measure the model’s performance.
For MMBench and SEED-Bench, the evaluation metric is accuracy. POPE’s performance is assessed using the F1 score, balancing precision and recall. MME scores are calculated using official scripts, with separate results for the Perception and Cognition tasks. Additionally, we report the normalized average score ($Avg^N$ \cite{Uniter-chen2020uniter,VALUE-li2021value}), facilitating fair comparisons across different tasks.

\subsection{Overall Results}
The results, as shown in \cref{tab:comp_others}, demonstrate that the MAGE model proposed in this study performs excellently in a 7B parameter scale LLM, particularly when using only 144 visual tokens. The MAGE model outperforms those using 256 and 576 visual tokens. For instance, MAGE scores 71.2 on the MMBench benchmark, significantly surpassing LLaVA-TokenPacker-HD (67.4) and LLaVA-1.5 (64.3). In MMBench-cn, MAGE scores 70.3, leading LLaVA-NeXT (62.3) and Qwen-VL-Chat (56.3). On the SEED benchmark, MAGE scores 65.3, surpassing C-Abstractor (64.5) and D-Abstractor (63.8).

Furthermore, MAGE 14B also excels on the POPE benchmarks, with a score of 89.9 on POPE, outperforming LLaVA-1.5 (85.9) and LLaVA-TokenPacker-HD (88.1), and scoring 73.9 on MMBench, surpassing LLaVA-TokenPacker-HD (69.5) and D-Abstractor (73.5). These results indicate that MAGE efficiently extracts visual features, significantly reduces computational costs while improving performance, demonstrating its competitiveness in multimodal tasks.

\subsection{Ablation Study}
\begin{table}[!t]
    \tablestyle{3pt}{1.05}
    \begin{center}
    \scalebox{0.95}{
                
        \begin{tabular}{@{}lcccccccccccccccccccc@{}}
    \toprule
    Model   &  \mmbdev & \mmbdevcn & \seedimg & {MME} & \pope  \\
    \midrule
    MAGE  &  \textbf{71.2} & \textbf{70.3} & \textbf{65.3} & \textbf{1572/323} & \textbf{86.7}\\
    \hspace{0.8em} w/o IAN proj.  &  68.1 & 67.8 & 62.9 & 1538/301 & 86.0 \\
    \hspace{0.8em} w/o align.  & 69.7 & 68.3 & 61.4 & 1541/307 & 86.2 \\
    \hspace{0.8em} w/o IAN \& align. & 67.0 & 65.8 & 60.9 & 1521/294 & 85.9 \\

  %   \midrule

  \bottomrule 
  \end{tabular}
    }
    \end{center}
    \caption{\small
        \textbf{Ablation study evaluating the impact of the IAN projector and alignment strategies in the proposed MAGE model.}
        The table demonstrates how the performance decreases when the IAN projector (w/o IAN proj.), the alignment strategy (w/o align.), or both (w/o IAN \& align.) are removed. 
    }
    \label{tab:ablation}
\end{table}

The impact of the IAN projection module (IAN proj.) and alignment strategy (align.) on model performance was evaluated through ablation experiments on five benchmark datasets: MMBench, MMBench-cn, SEED, MME, and POPE, with results summarized in \cref{tab:ablation}.

The full MAGE model achieved the best performance across all benchmarks. Removing the IAN projection module led to a performance drop, particularly on SEED (from 65.3 to 62.9). Further removal of the alignment strategy caused an additional decrease, with the SEED score dropping to 61.4. The lowest performance was observed when both components were removed, with the SEED score falling to 60.9. These results emphasize the critical role of the IAN module and alignment strategy in improving model performance.

\subsection{Analysis on IAN}
The experimental results, as shown in \cref{tab:spatial_task}, demonstrate that MAGE outperforms other designs across multiple tasks. In the MME POS task, MAGE scored 140, surpassing other bridging layers. In the MMBench SR task, MAGE achieved the best score of 39.2. Similarly, in the SEED-Bench SR task, MAGE ranked first with a score of 54.7 and an average score of 58.3, outperforming all other models.

Even with only 144 visual tokens, MAGE scored 33.3 in the MMBench SR task, outperforming models using 256 visual tokens, such as Resampler (22.2) and C-Abstractor (24.4), demonstrating its efficient reasoning under resource constraints.

Furthermore, the results in \cref{table:visual_token} show that increasing the number of visual tokens and model size leads to improved performance. In the MMB task, the 7B model improved from a score of 70.9 with 64 visual tokens to 71.4 with 256 visual tokens, while the 13B model improved from 73.1 to 74.4, indicating stable growth. The 13B model outperformed the 7B model in most tasks, suggesting that larger models with more visual tokens effectively enhance performance.

\begin{table}[!t]
    \tablestyle{3pt}{1.1}
    \begin{center}
    \scalebox{0.9}{
                
        \begin{tabular}{l|ll|cccccc|c}
        & \multirow{2}{*}{Projector} & \multirow{2}{*}{\#Token}  & MME & \multicolumn{3}{c}{MMB} & \multicolumn{2}{c|}{SEED} & \multirow{2}{*}{\avgn} \\ 
        & & &  POS & SR & OL & PR & SR & IL & \\ 
        \hline\hline

        B1 & Linear    & 576 &\textbf{140} & 24.4 & 40.7 & 70.8 & 48.9 & 60.9 & 52.6 \\
        
        \hline
        B2 &Resampler & 144  & 75.0 & 22.2 & 43.2 & 62.5 & 47.5 & 50.6 & 43.9 \\
        B3 &C-Abstractor    & 144  & 135 & 24.4 & \textbf{54.3} & 66.7 & 49.0 & \textbf{58.8} & 53.5 \\
        B4 &\ian     & 144  & 135 & \textbf{33.3} & 51.9 & \textbf{67.0} & \textbf{50.3} & 55.4 & \textbf{54.2} \\
        
        \hline
        B5 &Resampler & 256 & 73.3 & 24.4 & 37.0 & \textbf{79.2} & 44.4 & 51.8 & 45.6 \\
        B6 &C-Abstractor    & 256 & 136.7 & 26.7 & \textbf{55.6} & 75.0 & 52.7 & \textbf{59.3} & 56.3 \\
        B7 &\ian     & 256 & \textbf{140} & \textbf{39.2} & 53.1 & 76.5 & \textbf{54.7} & 56.1 & \textbf{58.3}
        
        \end{tabular}
    }
    \end{center}
    \caption{\small
        \textbf{Comparison of Different Projectors}
        The table presents results for various tasks like Position (POS), Spatial Relationship (SR), Object Localization (OL), and Physical Relation (PR) for MMBench, and SR and Instance Location (IL) for SEED-Bench. \avgn represents the normalized average across all six tasks. The number of visual tokens (\#Token) is also reported. Bold values indicate the best results.
    }
    \label{tab:spatial_task}
\end{table}

\begin{table}[!t]
    \tablestyle{3pt}{1.05}
    \begin{center}
    \scalebox{0.95}{
        \begin{tabular}{@{}cc|cccccc@{}}
        \toprule
        Model  & \#Token  & \mmbdev  &\mmbdevcn  & \mmep  & MME  & \seedimg & \pope   \\
        \midrule \midrule
        \multirow{3}{*}{7B}   & 64 & 70.9 & 70.0 & 1544 & 1851 & 64.8  & 86.3  \\
                    & 144   & 71.2 & 70.3 & 1572 & 1895 & 65.3  & 86.7  \\
                   & 256   & \textbf{71.4} & \textbf{70.6} & \textbf{1580} & \textbf{1913} & \textbf{65.6}  & \textbf{86.9}  \\
        \midrule
        \multirow{3}{*}{13B}   & 64 & 73.1   & 71.9 & 1595 & 1907 & 67.4 & 89.1   \\
                    & 144 & 73.9 & 73.0 & 1632 & 1944 & 67.8 & 89.9  \\
                    & 256 & \textbf{74.4} & \textbf{73.3} & \textbf{1643} & \textbf{1969} & \textbf{68.1} & \textbf{90.3}  \\
        \bottomrule
        \end{tabular}
    }
    \end{center}
    \caption{
        \small 
        \textbf{Ablation study on the effect of visual token settings. } The table compares two model sizes across different tokens (\#Token).
    }
    \label{table:visual_token}
\end{table}

\subsection{Tool Use Expand the Output Modality of the VLM}

\begin{figure}[!t]
    \begin{center}
    \scalebox{0.95}{
        \includegraphics[width=\linewidth]{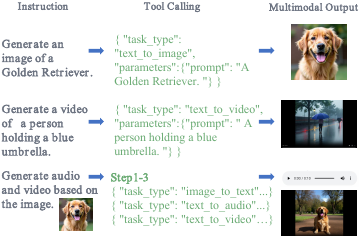}
    }
    \end{center}
    \caption{
         The figure shows an example of how our model extends multimodal applications by utilizing tools, including two basic tasks: text-to-image and text-to-video, as well as a composite task that generates images by combining both images and videos.
    }
    \label{fig:tool_use_examples}
\end{figure}
 Figure \ref{fig:tool_use_examples} illustrates three specific examples, covering two basic tasks: text-to-image and text-to-Video, as well as a more complex task: generating audio and video based on an uploaded image. In these tasks, the model automatically selects the most appropriate tool for processing based on the input text description or multimodal content, thereby generating high-quality outputs.

In particular, for situations where direct modality generation is not possible, the model coordinates the invocation of other tools for inference. As shown in case 3 in Figure \ref{fig:tool_use_examples}, when there is no available tool in the toolset to directly generate audio and video from an image, the model first calls the image-to-text tool to convert the image into a textual description, which is then used as the prompt for generating audio and video. This mechanism significantly enhances the adaptability and flexibility of VLMs when handling diverse tasks.
% Specifically, when there is no tool in the toolset that can directly generate audio and video from an image, as shown in case 3 of Figure \ref{fig:tool_use_examples}. This mechanism significantly enhances the adaptability and flexibility of visual language models in handling diverse tasks.
\section{Conclusion}

This study proposes MAGE (Multimodal Alignment and Generation Enhancement), a novel architecture that addresses semantic and dimensional gaps between visual encoders and large language models (LLMs). MAGE achieves efficient alignment through the Intelligent Alignment Network (IAN), optimizes generation and alignment using a dual-loss strategy, and supports flexible multimodal task outputs, such as image, audio, and video generation. Experimental results demonstrate that MAGE performs exceptionally well across various benchmarks, achieving state-of-the-art efficiency and accuracy.

\section*{Acknowledgements}
This research was supported by the Beijing Municipal Science and Technology Project and the GTCOM 2030 AI Research Institute (AIRIS).
\section*{Contribution Statement}
Shaojun E and Yuchen Yang contributed equally. Ziyan Chen is the corresponding author.
% \newpage
\bibliographystyle{named}
\bibliography{ijcai25}

\begin{thebibliography}{}

\bibitem[\protect\citeauthoryear{Alayrac \bgroup \em et al.\egroup }{2022}]{flamingo}
Jean-Baptiste Alayrac, Jeff Donahue, Pauline Luc, Antoine Miech, Iain Barr, Yana Hasson, Karel Lenc, Arthur Mensch, Katherine Millican, Malcolm Reynolds, et~al.
\newblock Flamingo: a visual language model for few-shot learning.
\newblock {\em Advances in neural information processing systems}, 35:23716--23736, 2022.

\bibitem[\protect\citeauthoryear{Anthropic}{2024}]{anthropic2024claude3d5}
Anthropic.
\newblock Claude 3.5 sonnet.
\newblock \url{https://www.anthropic.com/news/claude-3-5-sonnet}, 2024.
\newblock Accessed: 2024-7-21.

\bibitem[\protect\citeauthoryear{Bai \bgroup \em et al.\egroup }{2023}]{qwenvl-bai2023qwen}
Jinze Bai, Shuai Bai, Shusheng Yang, Shijie Wang, Sinan Tan, Peng Wang, Junyang Lin, Chang Zhou, and Jingren Zhou.
\newblock Qwen-vl: A versatile vision-language model for understanding, localization, text reading, and beyond.
\newblock {\em arXiv preprint arXiv:2308.12966}, 1(2):3, 2023.

\bibitem[\protect\citeauthoryear{Brown \bgroup \em et al.\egroup }{2020}]{brown2020language}
Tom Brown, Benjamin Mann, Nick Ryder, Melanie Subbiah, Jared~D Kaplan, Prafulla Dhariwal, Arvind Neelakantan, Pranav Shyam, Girish Sastry, Amanda Askell, et~al.
\newblock Language models are few-shot learners.
\newblock {\em Advances in neural information processing systems}, 33:1877--1901, 2020.

\bibitem[\protect\citeauthoryear{Cha \bgroup \em et al.\egroup }{2024}]{honeybee-cha2024honeybee}
Junbum Cha, Wooyoung Kang, Jonghwan Mun, and Byungseok Roh.
\newblock Honeybee: Locality-enhanced projector for multimodal llm.
\newblock In {\em Proceedings of the IEEE/CVF Conference on Computer Vision and Pattern Recognition}, pages 13817--13827, 2024.

\bibitem[\protect\citeauthoryear{Chen \bgroup \em et al.\egroup }{2020}]{Uniter-chen2020uniter}
Yen-Chun Chen, Linjie Li, Licheng Yu, Ahmed El~Kholy, Faisal Ahmed, Zhe Gan, Yu~Cheng, and Jingjing Liu.
\newblock Uniter: Universal image-text representation learning.
\newblock In {\em European conference on computer vision}, pages 104--120. Springer, 2020.

\bibitem[\protect\citeauthoryear{Chen \bgroup \em et al.\egroup }{2021}]{chen2021geoqa}
Jiaqi Chen, Jianheng Tang, Jinghui Qin, Xiaodan Liang, Lingbo Liu, Eric~P Xing, and Liang Lin.
\newblock Geoqa: A geometric question answering benchmark towards multimodal numerical reasoning.
\newblock {\em arXiv preprint arXiv:2105.14517}, 2021.

\bibitem[\protect\citeauthoryear{Chen \bgroup \em et al.\egroup }{2024}]{allava-chen2024allava}
Guiming~Hardy Chen, Shunian Chen, Ruifei Zhang, Junying Chen, Xiangbo Wu, Zhiyi Zhang, Zhihong Chen, Jianquan Li, Xiang Wan, and Benyou Wang.
\newblock Allava: Harnessing gpt4v-synthesized data for a lite vision-language model.
\newblock {\em arXiv preprint arXiv:2402.11684}, 2024.

\bibitem[\protect\citeauthoryear{Chiang \bgroup \em et al.\egroup }{2023}]{vicuna2023}
Wei-Lin Chiang, Zhuohan Li, Zi~Lin, Ying Sheng, Zhanghao Wu, Hao Zhang, Lianmin Zheng, Siyuan Zhuang, Yonghao Zhuang, Joseph~E. Gonzalez, Ion Stoica, and Eric~P. Xing.
\newblock Vicuna: An open-source chatbot impressing gpt-4 with 90\%* chatgpt quality, March 2023.

\bibitem[\protect\citeauthoryear{Dai \bgroup \em et al.\egroup }{2023}]{instructblip-dai2023instructblip}
Wenliang Dai, Junnan Li, Dongxu Li, Anthony Meng~Huat Tiong, Junqi Zhao, Weisheng Wang, Boyang Li, Pascale Fung, and Steven Hoi.
\newblock Instructblip: Towards general-purpose vision-language models with instruction tuning, 2023.

\bibitem[\protect\citeauthoryear{Dai \bgroup \em et al.\egroup }{2024}]{dai2024nvlm}
Wenliang Dai, Nayeon Lee, Boxin Wang, Zhuolin Yang, Zihan Liu, Jon Barker, Tuomas Rintamaki, Mohammad Shoeybi, Bryan Catanzaro, and Wei Ping.
\newblock Nvlm: Open frontier-class multimodal llms.
\newblock {\em arXiv preprint arXiv:2409.11402}, 2024.

\bibitem[\protect\citeauthoryear{Fu \bgroup \em et al.\egroup }{2024}]{mmedataset-fu2024mmecomprehensiveevaluationbenchmark}
Chaoyou Fu, Peixian Chen, Yunhang Shen, Yulei Qin, Mengdan Zhang, Xu~Lin, Jinrui Yang, Xiawu Zheng, Ke~Li, Xing Sun, Yunsheng Wu, and Rongrong Ji.
\newblock Mme: A comprehensive evaluation benchmark for multimodal large language models, 2024.

\bibitem[\protect\citeauthoryear{Hong \bgroup \em et al.\egroup }{2023}]{metagpt-hong2023metagpt}
Sirui Hong, Xiawu Zheng, Jonathan Chen, Yuheng Cheng, Jinlin Wang, Ceyao Zhang, Zili Wang, Steven Ka~Shing Yau, Zijuan Lin, Liyang Zhou, et~al.
\newblock Metagpt: Meta programming for multi-agent collaborative framework.
\newblock {\em arXiv preprint arXiv:2308.00352}, 2023.

\bibitem[\protect\citeauthoryear{Hu \bgroup \em et al.\egroup }{2021}]{lora-hu2021lora}
Edward~J Hu, Yelong Shen, Phillip Wallis, Zeyuan Allen-Zhu, Yuanzhi Li, Shean Wang, Lu~Wang, and Weizhu Chen.
\newblock Lora: Low-rank adaptation of large language models.
\newblock {\em arXiv preprint arXiv:2106.09685}, 2021.

\bibitem[\protect\citeauthoryear{Kafle \bgroup \em et al.\egroup }{2018}]{kafle2018dvqa}
Kushal Kafle, Brian Price, Scott Cohen, and Christopher Kanan.
\newblock Dvqa: Understanding data visualizations via question answering.
\newblock In {\em Proceedings of the IEEE conference on computer vision and pattern recognition}, pages 5648--5656, 2018.

\bibitem[\protect\citeauthoryear{Kim \bgroup \em et al.\egroup }{2022}]{kim2022ocr}
Geewook Kim, Teakgyu Hong, Moonbin Yim, JeongYeon Nam, Jinyoung Park, Jinyeong Yim, Wonseok Hwang, Sangdoo Yun, Dongyoon Han, and Seunghyun Park.
\newblock Ocr-free document understanding transformer.
\newblock In {\em European Conference on Computer Vision}, pages 498--517. Springer, 2022.

\bibitem[\protect\citeauthoryear{Li \bgroup \em et al.\egroup }{2021}]{VALUE-li2021value}
Linjie Li, Jie Lei, Zhe Gan, Licheng Yu, Yen-Chun Chen, Rohit Pillai, Yu~Cheng, Luowei Zhou, Xin~Eric Wang, William~Yang Wang, et~al.
\newblock Value: A multi-task benchmark for video-and-language understanding evaluation.
\newblock {\em arXiv preprint arXiv:2106.04632}, 2021.

\bibitem[\protect\citeauthoryear{Li \bgroup \em et al.\egroup }{2022}]{li2022blip}
Junnan Li, Dongxu Li, Caiming Xiong, and Steven Hoi.
\newblock Blip: Bootstrapping language-image pre-training for unified vision-language understanding and generation.
\newblock In {\em International conference on machine learning}, pages 12888--12900. PMLR, 2022.

\bibitem[\protect\citeauthoryear{Li \bgroup \em et al.\egroup }{2023a}]{seedbench-li2023seed}
Bohao Li, Rui Wang, Guangzhi Wang, Yuying Ge, Yixiao Ge, and Ying Shan.
\newblock Seed-bench: Benchmarking multimodal llms with generative comprehension.
\newblock {\em arXiv preprint arXiv:2307.16125}, 2023.

\bibitem[\protect\citeauthoryear{Li \bgroup \em et al.\egroup }{2023b}]{blip2-li2023blip}
Junnan Li, Dongxu Li, Silvio Savarese, and Steven Hoi.
\newblock Blip-2: Bootstrapping language-image pre-training with frozen image encoders and large language models.
\newblock In {\em International conference on machine learning}, pages 19730--19742. PMLR, 2023.

\bibitem[\protect\citeauthoryear{Li \bgroup \em et al.\egroup }{2023c}]{pope-li2023evaluating}
Yifan Li, Yifan Du, Kun Zhou, Jinpeng Wang, Wayne~Xin Zhao, and Ji-Rong Wen.
\newblock Evaluating object hallucination in large vision-language models.
\newblock {\em arXiv preprint arXiv:2305.10355}, 2023.

\bibitem[\protect\citeauthoryear{Li \bgroup \em et al.\egroup }{2024a}]{llavanext-li2024llava}
Feng Li, Renrui Zhang, Hao Zhang, Yuanhan Zhang, Bo~Li, Wei Li, Zejun Ma, and Chunyuan Li.
\newblock Llava-next-interleave: Tackling multi-image, video, and 3d in large multimodal models.
\newblock {\em arXiv preprint arXiv:2407.07895}, 2024.

\bibitem[\protect\citeauthoryear{Li \bgroup \em et al.\egroup }{2024b}]{tokenpacker-li2024tokenpacker}
Wentong Li, Yuqian Yuan, Jian Liu, Dongqi Tang, Song Wang, Jie Qin, Jianke Zhu, and Lei Zhang.
\newblock Tokenpacker: Efficient visual projector for multimodal llm.
\newblock {\em arXiv preprint arXiv:2407.02392}, 2024.

\bibitem[\protect\citeauthoryear{Liu \bgroup \em et al.\egroup }{2024a}]{llava1d5-liu2024improved}
Haotian Liu, Chunyuan Li, Yuheng Li, and Yong~Jae Lee.
\newblock Improved baselines with visual instruction tuning.
\newblock In {\em Proceedings of the IEEE/CVF Conference on Computer Vision and Pattern Recognition}, pages 26296--26306, 2024.

\bibitem[\protect\citeauthoryear{Liu \bgroup \em et al.\egroup }{2024b}]{llava-liu2024visual}
Haotian Liu, Chunyuan Li, Qingyang Wu, and Yong~Jae Lee.
\newblock Visual instruction tuning.
\newblock {\em Advances in neural information processing systems}, 36, 2024.

\bibitem[\protect\citeauthoryear{Liu \bgroup \em et al.\egroup }{2025a}]{llavaplus-liu2025llava}
Shilong Liu, Hao Cheng, Haotian Liu, Hao Zhang, Feng Li, Tianhe Ren, Xueyan Zou, Jianwei Yang, Hang Su, Jun Zhu, et~al.
\newblock Llava-plus: Learning to use tools for creating multimodal agents.
\newblock In {\em European Conference on Computer Vision}, pages 126--142. Springer, 2025.

\bibitem[\protect\citeauthoryear{Liu \bgroup \em et al.\egroup }{2025b}]{mmbench-liu2025mmbench}
Yuan Liu, Haodong Duan, Yuanhan Zhang, Bo~Li, Songyang Zhang, Wangbo Zhao, Yike Yuan, Jiaqi Wang, Conghui He, Ziwei Liu, et~al.
\newblock Mmbench: Is your multi-modal model an all-around player?
\newblock In {\em European Conference on Computer Vision}, pages 216--233. Springer, 2025.

\bibitem[\protect\citeauthoryear{Masry \bgroup \em et al.\egroup }{2022}]{masry2022chartqa}
Ahmed Masry, Do~Xuan Long, Jia~Qing Tan, Shafiq Joty, and Enamul Hoque.
\newblock Chartqa: A benchmark for question answering about charts with visual and logical reasoning.
\newblock {\em arXiv preprint arXiv:2203.10244}, 2022.

\bibitem[\protect\citeauthoryear{Mathew \bgroup \em et al.\egroup }{2021}]{mathew2021docvqa}
Minesh Mathew, Dimosthenis Karatzas, and CV~Jawahar.
\newblock Docvqa: A dataset for vqa on document images.
\newblock In {\em Proceedings of the IEEE/CVF winter conference on applications of computer vision}, pages 2200--2209, 2021.

\bibitem[\protect\citeauthoryear{OpenAI}{2024}]{openai2024hellogpt4o}
OpenAI.
\newblock Hello gpt-4o.
\newblock \url{https://openai.com/index/hello-gpt-4o/}, 2024.
\newblock Accessed: 2024-5-13.

\bibitem[\protect\citeauthoryear{Radford \bgroup \em et al.\egroup }{2021a}]{radford2021learning}
Alec Radford, Jong~Wook Kim, Chris Hallacy, Aditya Ramesh, Gabriel Goh, Sandhini Agarwal, Girish Sastry, Amanda Askell, Pamela Mishkin, Jack Clark, et~al.
\newblock Learning transferable visual models from natural language supervision.
\newblock In {\em International conference on machine learning}, pages 8748--8763. PMLR, 2021.

\bibitem[\protect\citeauthoryear{Radford \bgroup \em et al.\egroup }{2021b}]{clip-radford2021learning}
Alec Radford, Jong~Wook Kim, Chris Hallacy, Aditya Ramesh, Gabriel Goh, Sandhini Agarwal, Girish Sastry, Amanda Askell, Pamela Mishkin, Jack Clark, et~al.
\newblock Learning transferable visual models from natural language supervision.
\newblock In {\em International conference on machine learning}, pages 8748--8763. PMLR, 2021.

\bibitem[\protect\citeauthoryear{Shen \bgroup \em et al.\egroup }{2024}]{hugginggpt-shen2024hugginggpt}
Yongliang Shen, Kaitao Song, Xu~Tan, Dongsheng Li, Weiming Lu, and Yueting Zhuang.
\newblock Hugginggpt: Solving ai tasks with chatgpt and its friends in hugging face.
\newblock {\em Advances in Neural Information Processing Systems}, 36, 2024.

\bibitem[\protect\citeauthoryear{Team \bgroup \em et al.\egroup }{2023}]{team2023gemini}
Gemini Team, Rohan Anil, Sebastian Borgeaud, Jean-Baptiste Alayrac, Jiahui Yu, Radu Soricut, Johan Schalkwyk, Andrew~M Dai, Anja Hauth, Katie Millican, et~al.
\newblock Gemini: a family of highly capable multimodal models.
\newblock {\em arXiv preprint arXiv:2312.11805}, 2023.

\bibitem[\protect\citeauthoryear{Vaswani}{2017}]{vaswani2017attention}
A~Vaswani.
\newblock Attention is all you need.
\newblock {\em Advances in Neural Information Processing Systems}, 2017.

\bibitem[\protect\citeauthoryear{Wang \bgroup \em et al.\egroup }{2024}]{wang2024qwen2}
Peng Wang, Shuai Bai, Sinan Tan, Shijie Wang, Zhihao Fan, Jinze Bai, Keqin Chen, Xuejing Liu, Jialin Wang, Wenbin Ge, et~al.
\newblock Qwen2-vl: Enhancing vision-language model's perception of the world at any resolution.
\newblock {\em arXiv preprint arXiv:2409.12191}, 2024.

\bibitem[\protect\citeauthoryear{Wu}{2017}]{wu2017introduction}
Jianxin Wu.
\newblock Introduction to convolutional neural networks.
\newblock {\em National Key Lab for Novel Software Technology. Nanjing University. China}, 5(23):495, 2017.

\bibitem[\protect\citeauthoryear{Yao \bgroup \em et al.\egroup }{2024}]{yao2024deco}
Linli Yao, Lei Li, Shuhuai Ren, Lean Wang, Yuanxin Liu, Xu~Sun, and Lu~Hou.
\newblock Deco: Decoupling token compression from semantic abstraction in multimodal large language models.
\newblock {\em arXiv preprint arXiv:2405.20985}, 2024.

\bibitem[\protect\citeauthoryear{Yu \bgroup \em et al.\egroup }{2022}]{yu2022coca}
Jiahui Yu, Zirui Wang, Vijay Vasudevan, Legg Yeung, Mojtaba Seyedhosseini, and Yonghui Wu.
\newblock Coca: Contrastive captioners are image-text foundation models.
\newblock {\em arXiv preprint arXiv:2205.01917}, 2022.

\bibitem[\protect\citeauthoryear{Yuan \bgroup \em et al.\egroup }{2021}]{yuan2021florence}
Lu~Yuan, Dongdong Chen, Yi-Ling Chen, Noel Codella, Xiyang Dai, Jianfeng Gao, Houdong Hu, Xuedong Huang, Boxin Li, Chunyuan Li, et~al.
\newblock Florence: A new foundation model for computer vision.
\newblock {\em arXiv preprint arXiv:2111.11432}, 2021.

\bibitem[\protect\citeauthoryear{Zhao \bgroup \em et al.\egroup }{2024}]{zhao2024easygen}
Xiangyu Zhao, Bo~Liu, Qijiong Liu, Guangyuan Shi, and Xiao-Ming Wu.
\newblock Easygen: Easing multimodal generation with bidiffuser and llms.
\newblock In {\em Proceedings of the 62nd Annual Meeting of the Association for Computational Linguistics (Volume 1: Long Papers)}, pages 1351--1370, 2024.

\bibitem[\protect\citeauthoryear{Zheng \bgroup \em et al.\egroup }{2023}]{vicuna1d5-zheng2023judging}
Lianmin Zheng, Wei-Lin Chiang, Ying Sheng, Siyuan Zhuang, Zhanghao Wu, Yonghao Zhuang, Zi~Lin, Zhuohan Li, Dacheng Li, Eric Xing, et~al.
\newblock Judging llm-as-a-judge with mt-bench and chatbot arena.
\newblock {\em Advances in Neural Information Processing Systems}, 36:46595--46623, 2023.

\end{thebibliography}
% \newpage
% \input{sec/7_appendix}
\end{document}